# Cyclic Fusion of Measuring Information in Curved Elastomer Contact via Vision-Based Tactile Sensing

Zilan Li, Zhibin Zou, Weiliang Xu, Yuanzhi Zhou, Guoyuan Zhou, Muxing Huang, Xuan Huang, and Xinming Li, *Member, IEEE*

***Abstract*—Vision-based tactile sensors encode object data via optical signals, capturing microscale deformations using elastomer through densely arranged optical imaging sensors to detect subtle data variations. To enable continuous contact recognition, elastomers are crafted with curved surfaces to adjust to changes in the contact area. However, this design leads to uneven deformations, distorting tactile images and inaccurately reflecting the true elastomer deformations. In this work, we propose a cyclic fusion strategy for vision-based tactile sensing for precise contact data extraction and shape feature integration at the pixel level. Utilizing frequency domain fusion, the system merges topography as indicated by elastomer deformation, enhancing information content by 8%, and regional information bias is reduced by 20% when preserving structural consistency. Further, this system could effectively extract and summarize micro-scale contact features, decreasing erroneous predictions by 20% in defect detection via neural networks and reducing surface projection bias by 50% in surface depth reconstruction. Using this strategy, the measurement minimizes data interference, accurately depicting object morphology on tactile images and enhancing tactile sensation restoration.**

## I. INTRODUCTION

Tactile sensing [1]-[4] is significant in various domains, including robotics [5], electronic skin [6], and human-computer interaction [7]. To obtain sufficiently trustworthy tactile information in more complex scenarios, vision-based tactile sensors [8][9] have garnered substantial attention in precision manipulation [10], surface reconstruction [11], and related research endeavors, owing to their exceptional integration and impressive resolution. In the sensing system, the inbuilt camera achieves tactile reproduction by capturing the deformation produced by the elastomer in the sensor during interaction with the object. In continuous sensing, for planar elastomers, if the position of the contact needs to be changed, it is necessary to disengage from the object and then recontact it. This process leads to discontinuities in the acquisition. To overcome this difficulty, the elastomer can be arranged as a curved surface. The curved structure allows the elastomer to retain some areas out of contact and can be rolled to actively bring different locations in the structure into contact with the object, keeping the sensing system in contact at all times. This structural feature introduces the problem that the precise imaging plane of the camera does not match the curved surface of the contact structure. The accuracy of converting deformation data to images depends on the deformation occurring within the camera's effective focal length. However, in systems with curved contact structures, this condition isn't consistently met, disrupting the correlation between captured images and contact data. In addition, the measuring information cannot directly represent the actual morphology of the object due to the inconsistency of the contact morphology in the curved structures in each region.

Faced with this situation, the concern is to achieve a truthful recording and identification of the deformation variables of elastomers arranged on curved surfaces. An approach is to accurately align the system design and component arrangement to enable conversion from elastomer deformation to tactile images being accurately recognized. For example, the TacTip sensor has a finger-like shape to achieve effective contact over a multi-angle area [12]. Achieving congruence with the contact surface often necessitates a complex and precise sensing system design to ensure comprehensive coverage of the intended measuring area. Typically, these approaches have a clear and fixed size requirement of the sensing target. Another approach is to select a certain range of tactile data from the image and stitch it according to the motion strategy. In related studies, the method of data consolidation relies on the aggregation of small, centrally located, motion-captured image segments presumed to be at a uniform depth [13]. This method demands precise temporal synchronization and fails to harness tactile information across other image areas efficiently. Although curved structures provide structural feasibility for continuous detection scenarios, including objects larger than the sensor size or continuous non-stop production lines, this curvature challenges the consistency of captured contact depth, and the camera's focal plane, typically aligned with the surface, can result in out-of-focus imaging. This complicates the consistency and usability of the data. It is of concern that the elastomer deformation information and contact structure

This work was supported by the Guangdong Basic and Applied Basic Research Foundation (No. 2022A1515010136), the Guangdong Provincial Key Laboratory of Nanophotonic Functional Materials and Devices, and the South China Normal University start-up fund.

Zilan Li, Zhibin Zou, Weiliang Xu, Yuanzhi Zhou, Guoyuan Zhou, Muxing Huang, Xuan Huang, and Xinming Li are with the Guangdong Provincial Key Laboratory of Nanophotonic Functional Materials and Devices, Guangdong Basic Research Center of Excellence for Structure and Fundamental Interactions of Matter, School of Optoelectronic Science and Engineering, South China Normal University, Guangzhou 510006, China (corresponding author e-mail: xmli@m.scnu.edu.cn).

information contained in the acquired tactile data should be identifiable and separable in a defined sensing system. Meanwhile, on a series of tactile images obtained during the acquisition process, the actual tactile information can be fused by analyzing and filtering the features between the images.

To synthesize elastomer deformation information and overcome the interference of inconsistent contact depths, we propose a cyclic fusion strategy for vision-based tactile sensing. In the fusion strategy, the stitching of actual contact data is realized by fusing tactile information from multiple images. Considering these images encapsulate diverse contact states at varying depths and are distributed across an image sequence, the strategy extracts the overall features and edge features of the object's contact in the frequency domain. Further, different features in the tactile image are taken for different fusion computations around the contact depth distribution feature of the surface structure to retain the effective part of it. By employing the fusion strategy, pixel-level integration of tactile images resulted in an 8% improvement in information content while preserving structural similarity. Additionally, regional information bias was effectively reduced by 20%. In neural network-based defect detection applications, the use of this fusion approach led to a 20% reduction in erroneous predictions. Furthermore, when applied to surface depth reconstruction, the strategy significantly reduced the projection bias resulting from surface curvature by 50%. These results indicate the robustness and adaptability of the proposed method in enhancing tactile image quality and improving application-specific outcomes.

With the proposed cyclic fusion strategy, it can be achieved to remove the data interference from the contact structure for the successive acquisition of tactile information thereby fusing and highlighting the information representing the actual morphology of the object on a tactile image, which facilitates the restoration of tactile sensations. The ability of the fusion strategy on the scale of a hundred micrometers is verified using tactile information at the pixel level in tactile images. The main contributions of this work include:

1) Information fusion based on the data backward and forward relationships of the contact sensing process allows pixel-level processing on tactile images and provides the amount of information available for images.
2) No complex a priori knowledge and a wide range of applicability are required. The fact that the fusion process has no requirements on the structural morphology of the surface and the number of fusions is optional means that the strategy is executed without the inclusion of additional system parameters. Moreover, this strategy applies to both image-based RGB fusion and surface reconstruction via photometric stereo.

In Table I the ways of performing tactile data correction and reduction on the curved surface are compared. Our salient advantages compared to other schemes are the absence of a priori system parameters involved in the calibration and the simplicity of the computational approach, as well as the generality of its use in different contexts.

TABLE I

COMPARISON OF METHODS FOR TACTILE DATA CORRECTION AND REDUCTION ON CURVED SURFACES

| Methods | Advantages | Disadvantages |
|---|---|---|
| Time-splicing based on motion [13] | Simplicity of implementation<br>Can be considered consistent depth | Only linear motion<br>Poor robustness of motion<br>Low data utilization |
| Parameter correction based on structure [12] | High precision<br>Less data processing | Complex structural design<br>Lack of some generalization |
| Data reduction based on deep learning [11] | High degree of automation<br>Ability to fit complex relationships | High demand for data |
| Cyclic fusion based on sequence(this work) | Not requiring prior knowledge<br>Simple data requirements | Requires data relevance |

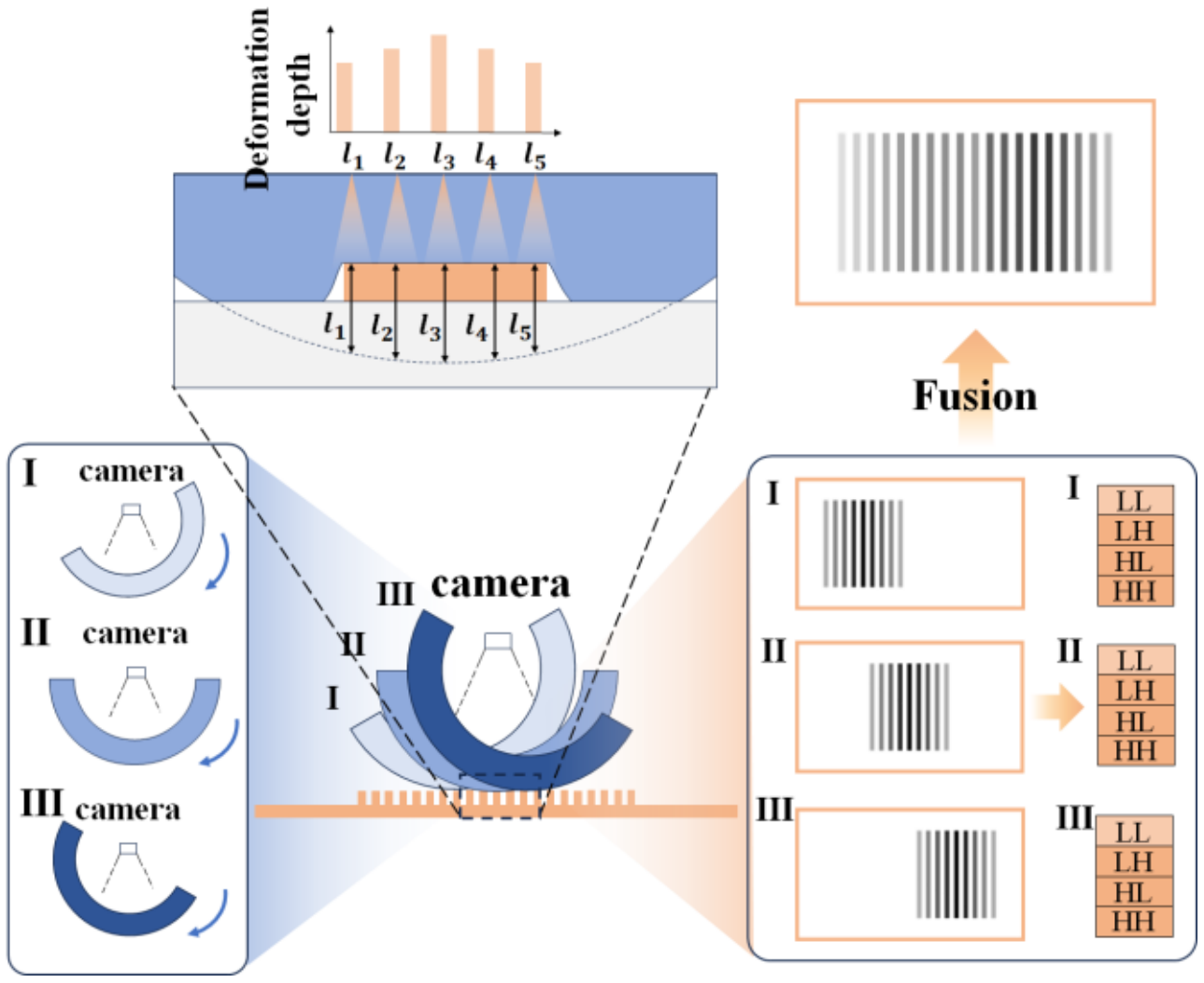

**Fig. 1.** The principle of the proposed cycle fusion strategy. Curved surface structure permits continuous contact and acquisition of the object while the sensor is rolling. During a single acquisition, due to the characteristics of the curved surface contact structure, the deformations generated at different positions of the sensor are inconsistent. The effect of fusion and the main processes of the fusion strategy are demonstrated in a distribution of simulated data.

## II. PRINCIPLES AND REASONS FOR FUSION

### *A. Principles of cyclic fusion strategy*

Despite advanced processing techniques like multi-scale analysis [14], and wavefront optimization [15], each with unique advantages for image fusion, the fusion of contact surface data encompassing information transmission mechanisms and contact dynamics, remains underexplored. By filtering the actual elastomeric shape variables from continuously collected tactile data, combined with the high resolution of tactile images, the integration of object contact samples in pixel fusion can be achieved and theoretically generalized to similar surface contacts. Fig.1 illustrates the principle of the proposed cycle fusion strategy. Tactile information about different locations of the object is captured by scrolling, they are contact captures of different locations of the object and also varying contact depths for tactile information present in the same image. The sensing unit in the camera transforms the morphometric variables on tiny areas and presents them as pixel values. The strategy summarizes the information that reflects the actual shape of the contact location in the image and presents it in a single image. As shown in Fig. 2(a), the images involved in the fusion are taken from the frame images of the continuous detection process, and the distributed tactile information is integrated into a separate image during the fusion process to make a more complete record of the contact process. This process strictly adheres to the relative relationship of positions, even though they originally existed on separate images. This approach was applied in minor deformations, capturing the texture of a 0.1 mm bump in a resolution measurement board as a tactile image, illustrated in Fig.2(b).

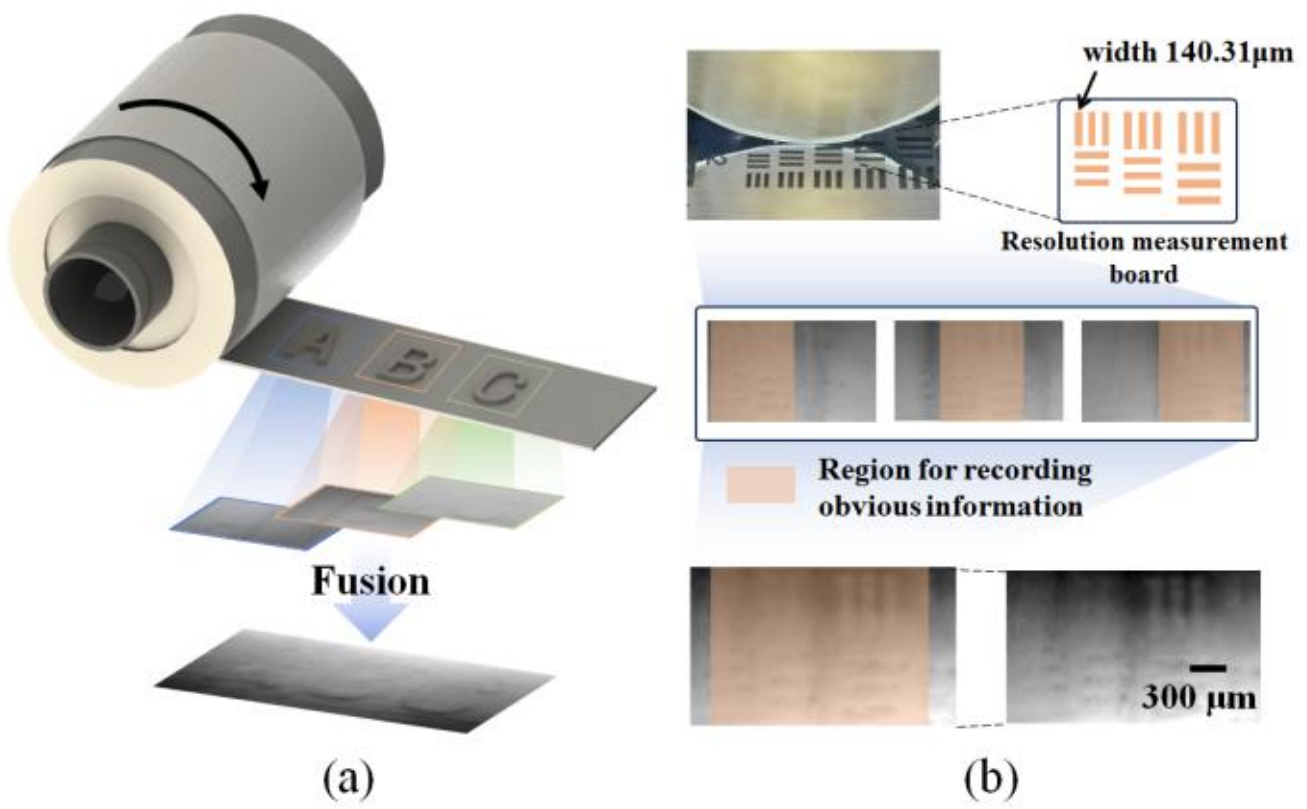

**Fig. 2.** (a) The fused data comes from frame image sampling in continuous detection, and the fusion allows distributed information to be integrated into the same image. (b) Data acquisition and fusion were performed at the resolution measurement board.

*B. Contribution of curved surface structures to tactile images*

In an acquisition sequence, distinct deformations are reflected in different regions of the image. The fusion strategy facilitates the integration of deformation data transmitted through the elastomer, allowing precise merging at the micrometer scale. To practically conduct the experimental discussion, the cylindrical vision-based tactile sensing system specifying the elastic contact structure of the curved surface as the region covering the outer surface of the cylinder was fabricated. This structure has been reported to be utilized in detection tasks [16][17]. The PDMS [18] is used to make elastomers assuming the functions of light transmission and flexibility. A discussion of the sensing system's structure is presented in Supplementary Material 1.

In a vision-based tactile sensor, the images hold measuring information captured by the camera, reflecting changes in the flexible contact layer via optical signals. The encoding of contact deformation by optical information can be analyzed using Lambert's cosine law [19] and is discussed in the experiments in Supplementary Material 2. The geometric relationships of the cylindrical structure are explored in Supplementary Material 3. From the whole image, the deformation induced by an object in contact with the sensor is not uniform for each region of the flexible contact layer due to the cylindrical structure. This leads to the fact that the depth of deformation produced by different sensor locations in contact with the same object is also inconsistent. As the sensor rolls, different sensor regions come into contact with the object, as shown in Fig.3(a). During continuous detection, the contact with an object is sustained for a period of time, during which the deformation pattern produced by the object is translated over the frame image and the depth of contact is constantly changing. This results in the acquired tactile images containing deformation data from different contact depths. As the sensor rolls, the contact process progresses from edge to center and then back to the edge, effectively capturing contact information at various depths. It starts from a shallow depth, progresses to a deeper level, and then returns to a shallower depth within a localized area.

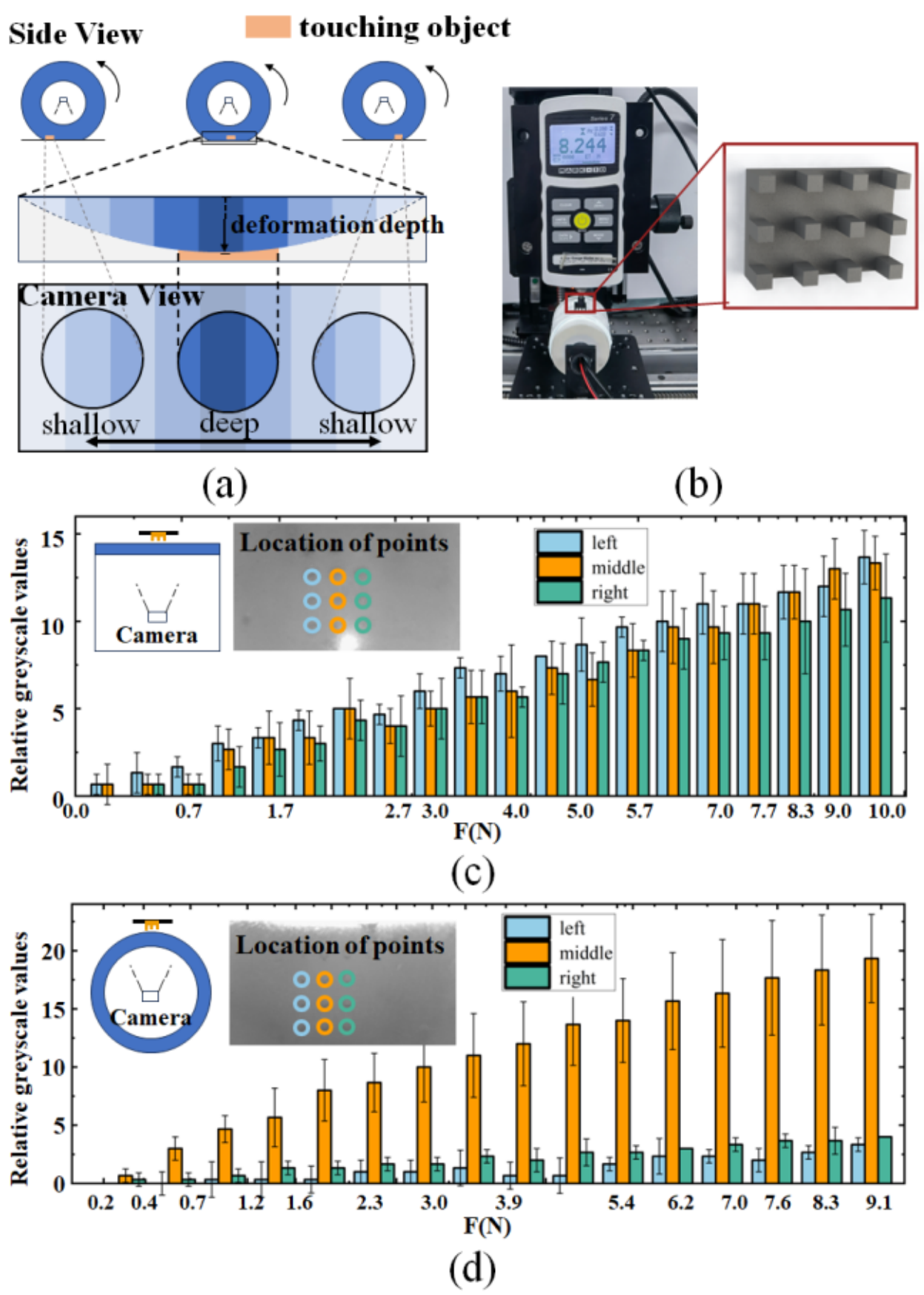

**Fig. 3.** (a) Different positions of the sensor during contact with the object correspond to various depths of data. (b) The scenario of the test. Matrix probes are used to simultaneously generate contact responses at various locations on the image. (c)(d) The schematic diagrams of the testing of the planar and curve structures, the location of the sampling point, and the relationship between the magnitude of the normal force and the amount of change in pixel value when pressed at different locations.

The cylindrical vision-based tactile sensor structure is distinct in its image content compared to planar sensors which is made notable in the comparison. This means that the appearance of the object is contained at various locations in the acquired images within a region. The vision-based tactile sensor for contrasting planar contact areas was prepared using the same materials and processes. The sensor was placed on the test platform employing a force gauge and displacement stage for testing, shown in Fig.3(b). A 3D-printed matrix probe, composed of 4x3 rectangular structures (1 mm x 1 mm x 2 mm) with 2 mm spacing, was mounted on the force gauge to evaluate deformation characteristics. Deformational characteristics were indicated by measuring the difference in gray values between images taken at sampling locations and images captured when no force was applied. Due to the presence of three points that did not produce a significant deformation response in the curved surface structure, three equidistant sampling points were chosen on each side of the contact area. In Fig.3(c), we observe that the planar structure exhibits uniform deformation across its regions. In contrast, Fig.3(d) shows that the change in gray value in the middle of the cylindrical structure is significantly higher than the sides. This shows that the cylindrical structure causes the center region to have a greater contact depth for the same force, resulting in greater deformation. The regions on the sides show a gradual response as the force increases. This results in the acquired tactile images containing deformation data from different contact depths. Each tactile image records contact data at various depths within a region. And there is a mixture of valuable data and disturbing data. Therefore, the selection and integration of contact data in the image sequence that can construct the actual appearance of the object is a critical part of surface contact data processing.

## III. Fusion using image sequences

### *A. Processes and methods of fusion*

Analyzing contact depth disparities attributed to surface structure nuances underscores the imperative for a streamlined and potent data processing approach. From the analysis, it can be obtained that there is potentially complementary information, despite the displacement of the objects and the non-corresponding presentation of the contact data in each image. Analyzing contact depth disparities attributed to surface structure nuances underscores the imperative for a streamlined and potent data processing approach. When this information is effectively integrated, it will provide a comprehensive and accurate description of the contact state. Therefore, we propose a cyclic fusion strategy. This is achieved through cyclic extraction and amalgamation of preceding and succeeding images in the sequence.

Since the target has a certain projected area, the image sequence containing the target exhibits distributed contact data from these deepest positions after rolling acquisition by the sensor. The task is to extract and fuse the contact data containing the target object. In an image sequence, the whole shape of the object and the edge information are present at different frequencies, which can be used to identify the object's actual shape, so it is important to separate them from the background and the noise. Confronted with this necessity, Daubechies wavelet transform [20] becomes an important tool. This transform can decompose an image at multiple resolutions so that the important frequency components can be separated, thus preserving and integrating the key contact features of the object while suppressing the noise.

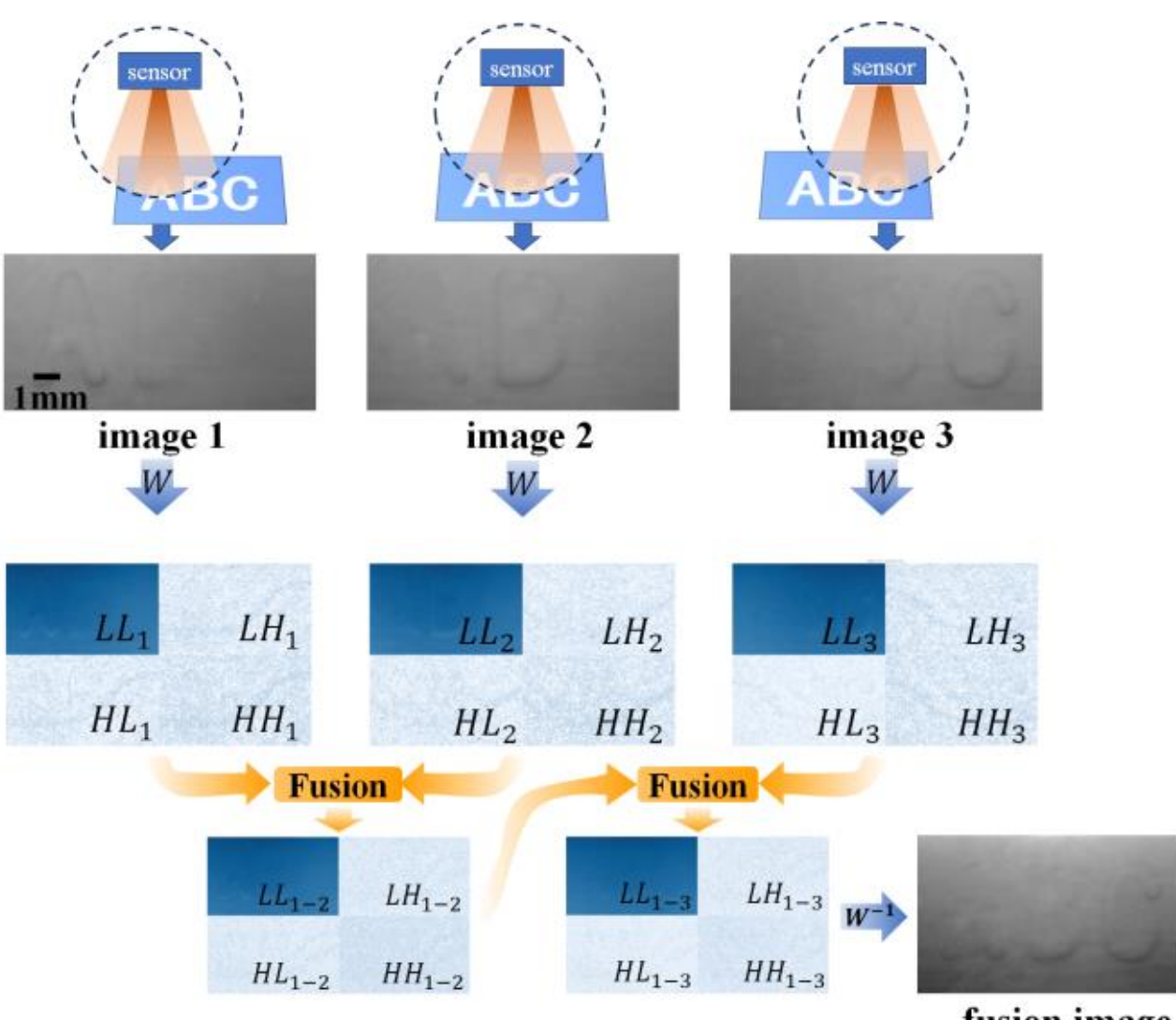

**Fig. 4.** The execution of the cyclic fusion process. From top to bottom, it demonstrates the sensor rolling acquisition of images at different locations, the aligned single frame image, the subband visualization map obtained by performing wavelet transform on the image, the cyclic fusion process, and the fused image. The 1mm scale in the figure applies to the captured image.

Fig. 4 illustrates the process of fusion over a sequence of letter images. The sensor passes over the top of each letter in this way during the successive detection process, where the letter at that position can have the deepest contact depth and acquire a more realistic topography. For acquiring frame images of different letters, the template-matching algorithm [21] was used to correct the image alignment error caused by the sensor shifts, to make the spatial distribution of the object features in the fused images as consistent as possible. Wavelet transform is applied to the image to obtain:

$$W(I) = \{LL, LH, HL, HH\} \tag{1}$$

where $W(I)$ refers to performing wavelet transform on image $I$ . $LL$ is the low-frequency component, which is used to capture the basic structure and global features of the object. $LH, HL$ and $HH$ are the three high-frequency components. They are computed in the horizontal, vertical, and diagonal directions, respectively, and carry subtle features such as edges and details. Applying wavelet transform allows us to independently distinguish and process different frequency components of an image. Meanwhile, the wavelet transform is highly sensitive to sharp regions in the image, usually the edges of objects and texture mutations, which means that the features in these regions will be extracted efficiently, and thus more easily recognized and preserved in the subsequent fusion process, ultimately retaining as much visually important information as possible.

Various fusion strategies are used for the subbands to cope with different features. The low-frequency subbands are fused equally, while the high-frequency subbands are assigned different weights for fusion after a significance analysis. With different fusion strategies, on the one hand, averaging the low-frequency subbands ensures the consistency of the fused images in terms of visual continuity and scene structure based on random noise reduction. On the other hand, the clarity of detail in the image is strengthened by performing saliency analysis and weighted fusion on the high-frequency subbands. The saliency analysis ensures that the fusion process pays more attention to the regions that are more significant for the comprehension of the scene, which presents the more informative and visually essential parts of the image. The fusion process is represented as:

$$F_{LL} = \frac{LL_1 + LL_2}{2} \tag{2}$$

$$Sal(I) = \left( \frac{\left| \frac{\partial^2 I}{\partial x^2} + \frac{\partial^2 I}{\partial y^2} \right|}{max\left( \left| \frac{\partial^2 I}{\partial x^2} + \frac{\partial^2 I}{\partial y^2} \right| \right)} \right) \times 255 \tag{3}$$

$$S = exp\left( \frac{Sal(I)}{255} \right) - 1 \tag{4}$$

$$W_1 = \frac{S_1}{S_1 + S_2},\ W_2 = 1 - W_1 \tag{5}$$

$$F_{HS} = HS_1 \cdot W_1 + HS_2 \cdot W_2, \quad HS = \{LH, HL, HH\} \tag{6}$$

where $F_{LL}$ denotes the fused low-frequency subband and $LL_1, LL_2$ denotes the low-frequency subband of the two images involved in the fusion. $Sal(I)$ represents the saliency map of an image, represented through second-order derivation of the image and transformed to a range of grey values. $S$ is the adjusted saliency map, which aims to optimize the distribution of salient components. Further, the weights $W_1$ and $W_2$ in the fusion can be obtained from the saliency maps $S_1$ and $S_2$ of the two images and $F_{HS}$ represents the fused high-frequency subbands with three specific results $LH$, $HL$ and $HH$. The fused subbands are converted back to the spatial domain by wavelet inverse transformation to get the fused image:

$$I_f = W^{-1}(F_{LL}, F_{LH}, F_{HL}, F_{HH}) \tag{7}$$

where $W^{-1}$ denotes the wavelet inverse transform. $I_f$ is the fused data remapped to grayscale. Afterward, the fused data is fused with the next image in the sequence as a new image by cyclic fusion strategy, and the process is repeated until completion. The cyclic fusion strategy is employed to fully utilize the effective information provided by each frame in the sequence to build up a complete and detail-rich image gradually. By performing the same transformation and fusion process of the fused image with the next image in the sequence, the detailed information on the critical areas can be gradually enhanced in each step and ensure that the fusion result maintains visual and informational coherence throughout the sequence. The fused image takes the actual measuring information (clear alphabets seen in the image), which initially existed in separate images, and fused them on a single image. This shows that our method realizes the splicing of effective information in the contact process.

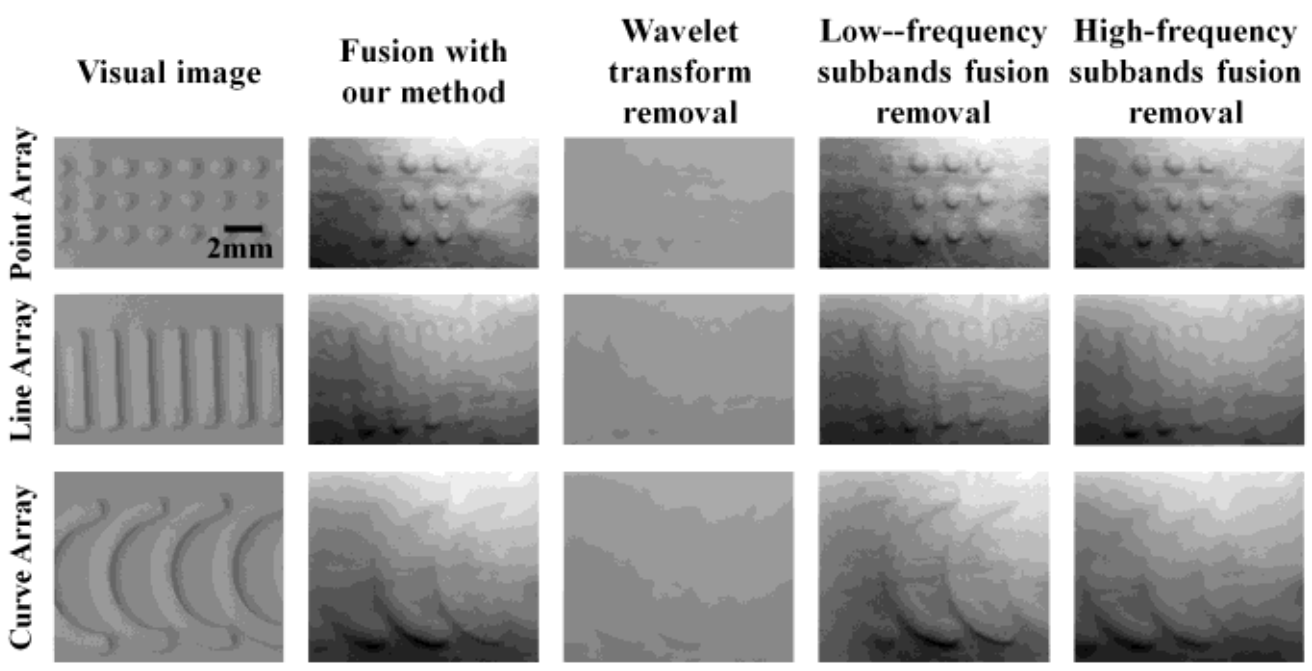

**Fig. 5.** Results of the ablation experiment. Shows visual images captured by the camera, fusion images using our method, and the fusion images obtained from the three sets of ablation experiments from left to right for the three pattern samples. A 2 mm width label is marked out in the tactile image on top, which applies to all images in this figure.

TABLE II

RESULTS OF ABLATION EXPERIMENTS

| Motif | Metrics | Fusion with our method | Wavelet Transform Removal | Low-frequency Subbands Fusion Removal | High-frequency Subbands Fusion Removal |
|---|---|---|---|---|---|
| Point | IE | 7.5814 | 4.9107 | 7.6178 | 7.5047 |
| | MS-SSIM | 0.4496 | 0.6166 | 0.4231 | 0.4500 |
| Line | IE | 7.5077 | 5.2018 | 7.4953 | 7.4977 |
| | MS-SSIM | 0.6407 | 0.7329 | 0.6180 | 0.6367 |
| Curve | IE | 7.6736 | 5.4270 | 7.5711 | 7.6485 |
| | MS-SSIM | 0.4947 | 0.5925 | 0.5010 | 0.4885 |

*B. Effectiveness evaluation of fusion*

The idea of ablation experiments [22] was adopted for validation to evaluate the effectiveness of the proposed image fusion strategy. By systematically removing the key steps in the fusion process, it aims to reveal the contribution of each part of the fusion strategy to the final fusion effect. Three main operations were performed ablation: the application of wavelet transform, the averaging fusion of the low-frequency subbands, and the fusion of the high-frequency subbands with weights assigned based on significance analysis. Specifically, the value of the wavelet transform was first compared by performing a direct averaging calculation of the pixels instead of applying the wavelet transform. Second, the contribution of the low-frequency part of the fusion is verified by selecting the low-frequency subbands of one image instead of the mean fusion of the low-frequency subbands of all images of the image sequence. Lastly, the weights of the adaptive high-frequency information obtained utilizing significance analysis are ignored and substituted for the information selected as the highest value. This idea typically means that the most salient portion of each section represents the sensitive information that receives the most attention.

For comparison, standardized samples were designed and produced to validate the proposed strategy. An image matrix elevated 1 mm from the plane was created for the experiment, with points, lines, and curves as pattern units. Two metrics were used to evaluate the performance of the proposed strategy: Information Entropy (IE) [23] and Multi-Scale Structural Similarity Index (MS-SSIM) [24]. IE measures the information richness of an image, which is the uncertainty in the distribution of image pixels, and a high value of information entropy indicates that the image contains more information. The purpose of choosing IE as a metric is to directly assess the amount of information in the fused image and determine whether the fusion operation effectively enhances the image content. MS-SSIM is used to quantify the visual similarity between the fused image and a selected reference image to assess the ability of the fused image to maintain the original visual information and structure. An optical image of a standardized sample taken directly through the camera as the reference image and scaled to a consistent scale. Scaling to the same scale and distributing the gray values to a uniform range allows the results to be focused on the evaluation of the degree of structural reproduction and reduces the possibility of interference by other factors such as average brightness. MS-SSIM was chosen as the evaluation metric because it reflects the maintenance and enhancement of image quality during the fusion process. Thus, the combination of MS-SSIM and IE provides a comprehensive framework for assessing both the visual accuracy of the fused images and measuring the increase in their information content, which is essential for validating the effectiveness of image fusion methods. The fusion process uses four images obtained from an acquisition. The results are shown in Fig. 5 and Table. II.

The results show that after removing one step, the majority of the IE values show a decrease. In particular, the fusion without wavelet transform fails to present the primary information of the images, both in terms of metrics and visual effects. In the case of MS-SSIM, the fusion without wavelet transform achieves high results because it averages the pixel distribution area in the image, which is similar to the effect in the visual image. The other results are relatively similar. The strategy's overall quantitative metrics and visual results are more balanced than the fusion strategy which removes one step. Meanwhile, the fusion strategy is relatively more robust for pattern features with different complexity, such as points and curves in the experiment whose feature complexity differs. In addition, the strategy also performs more consistently in removing noise and highlighting features on visual images. For example, there is blurring of the edges of the pattern in the image with the removal of the high-frequency subbands fusion strategy.

The effectiveness of the strategy was evaluated through comparison experiments. A sequence of images was captured using the sensor and a different number of images were for fusion. Since the motion of the sensor is characterized by translation, the fusion of tactile information is also characterized by a significant cumulative effect in the width direction, which is an essential feature of the proposed strategy on tactile images. Therefore, evaluating and calculating the deviation by region can reflect the effect of fusion. The adopted evaluation metrics IE and MS-SSIM were calculated on the whole image as well as on each of the 10 homogeneous intervals divided along the width direction, and the standard deviation was calculated according to the following equation:

$$\sigma_{IE} = \sqrt{\frac{\sum_{n=1}^{10}(IE_n - \overline{IE})^2}{10}} \tag{8}$$

$$\sigma_{MS-SSIM} = \sqrt{\frac{\sum_{n=1}^{10}(MS-SSIM_n - \overline{MS-SSIM})^2}{10}} \tag{9}$$

where $IE_n, MS-SSIM_n, n = 1,2 \ldots 10$ refers to the results obtained on each interval, $\overline{IE}$ and $\overline{MS-SSIM}$ is the average of these 10 results, and the calculated values obtained for $\sigma_{IE}$ and $\sigma_{MS-SSIM}$ are presented as "IE-std" and "MS-SSIM-std" in TABLE III.

For contrast, single images from the sequence and images using the sensor motion-based time sampling stitching method were also tested. The images obtained by time splicing are weighted and fused at the seams to avoid the creation of mutated regions. Due to the different distribution of information in different fusion strategies, the images undergoing evaluation are normalized for gray values to unify the mapping relationships, as shown in Fig. 6 and Table III. All the elements in the single images from the sequence and images using the sensor motion-based time sampling stitching method are directly derived from the acquired raw data and thus have a natural advantage in terms of similarity. Our proposed cyclic fusion strategies significantly improve the IE

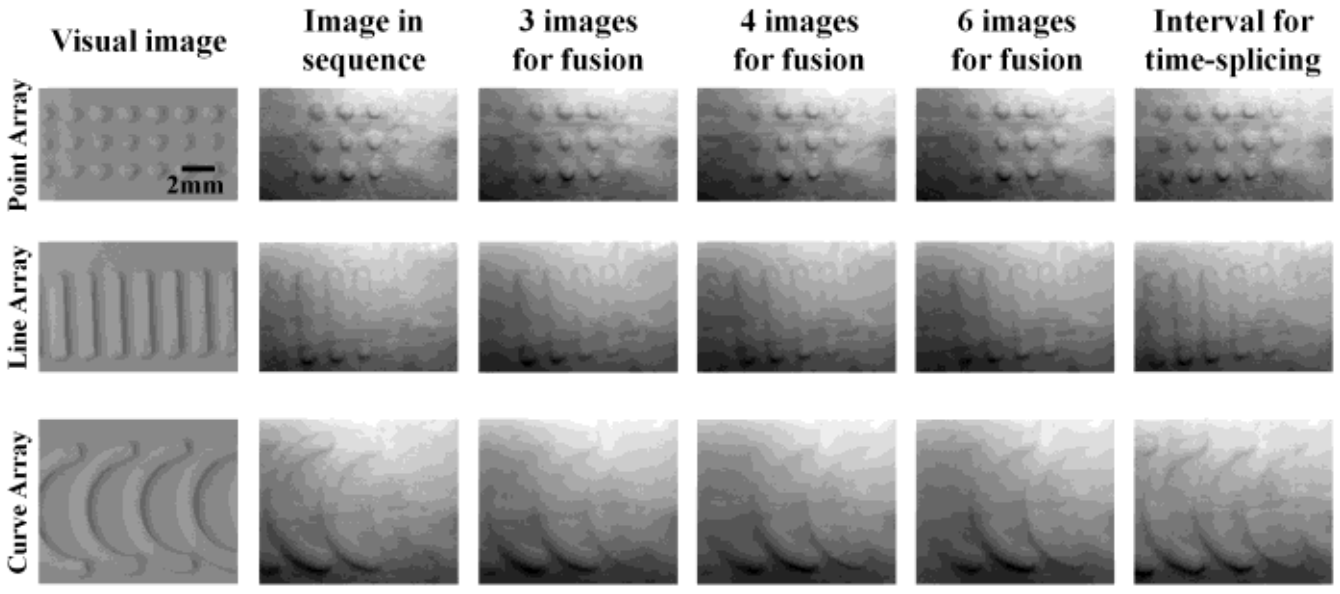

**Fig. 6.** Test results in the graphic matrices. Shows the original image, different numbers of fusion results, and the splicing performed using the sensor motion-based time sampling stitching method. A 2 mm width label is marked out in the tactile image on top, which applies to all images in this figure.

TABLE III

RESULTS OF COMPARISON OF METHODS AND QUANTITATIVE STRATEGIES

| Motif | Metrics | Single image | Fusion 3 images | Fusion 4 images | Fusion 6 images | Time-splicing |
|---|---|---|---|---|---|---|
| Point | IE | 6.7763 | 7.5111 | 7.5814 | 7.6193 | 6.9908 |
| | IE_std | 0.3692 | 0.2168 | 0.2228 | 0.2206 | 0.2872 |
| | MS-SSIM | 0.4653 | 0.4583 | 0.4496 | 0.4427 | 0.4727 |
| | MS-SSIM_std | 0.0495 | 0.0522 | 0.0537 | 0.0553 | 0.0504 |
| Line | IE | 7.0609 | 7.5068 | 7.5077 | 7.5293 | 7.0870 |
| | IE_std | 0.1354 | 0.1029 | 0.0994 | 0.0991 | 0.1258 |
| | MS-SSIM | 0.6351 | 0.6375 | 0.6407 | 0.6348 | 0.6312 |
| | MS-SSIM_std | 0.0826 | 0.0785 | 0.0775 | 0.0782 | 0.0807 |
| Curve | IE | 7.0938 | 7.6912 | 7.6736 | 7.6495 | 7.1128 |
| | IE_std | 0.1369 | 0.1095 | 0.1455 | 0.1728 | 0.3140 |
| | MS-SSIM | 0.5056 | 0.4931 | 0.4947 | 0.4993 | 0.5027 |
| | MS-SSIM_std | 0.0260 | 0.0259 | 0.0252 | 0.0266 | 0.0224 |

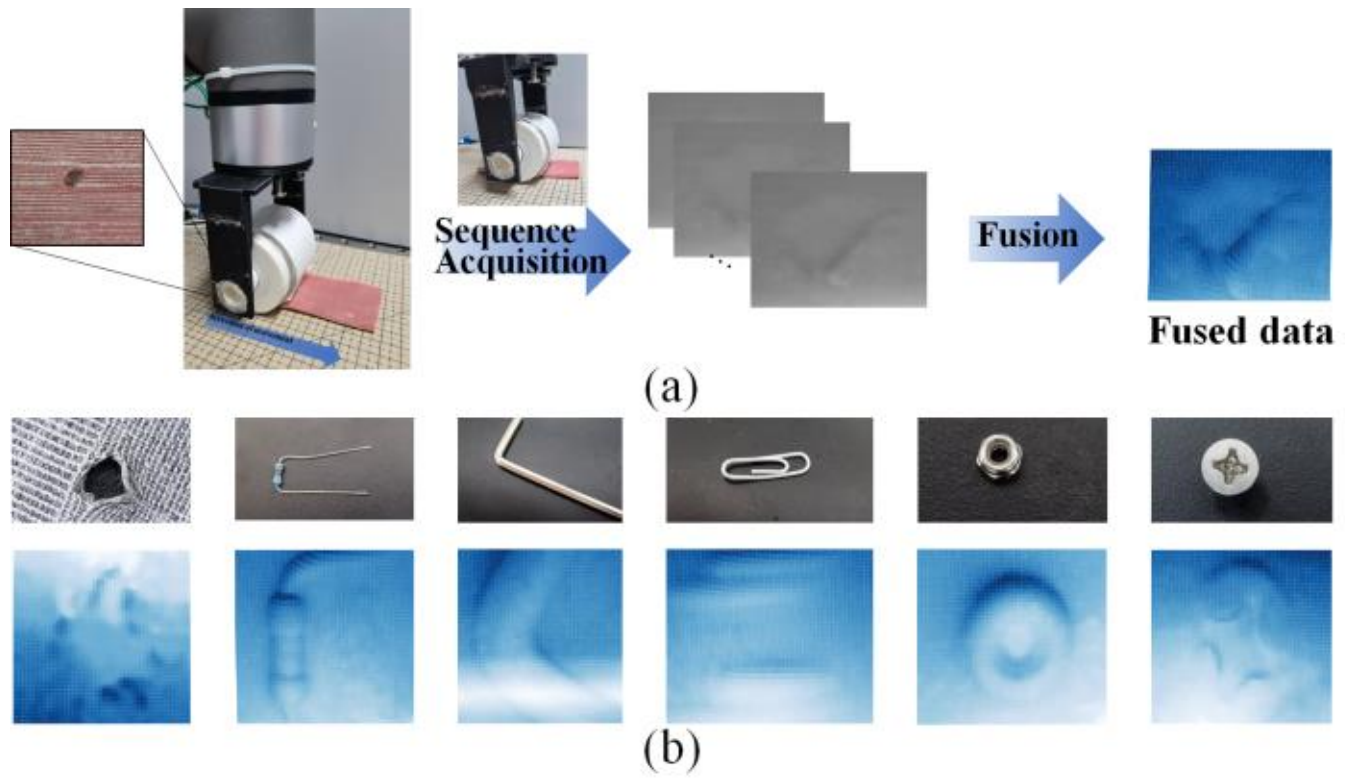

**Fig. 7.** (a)Sensors are used in combination with robotic arms for the process and results of object contact acquisition. (b) Application demonstrations on daily objects.

values while maintaining the same MS-SSIM values as these images. As can be seen from the regional differences embodied in the standard deviation, the use of the cyclic fusion strategy results in a uniform distribution of the amount of information at each location of the image, which implies that the tactile information captured on different frames is uniformly collated at each location of an image. In pixel fusion, the amount of information is increased by 8% and the regional deviation of information is reduced by 20% while maintaining consistent structural similarity. The significant advantage of the IE standard deviation also illustrates that the proposed strategy effectively mitigates the problem of information dispersion due to depth inconsistency in curved surface contact. Although the time splicing method is also able to collect these contact data with approximately consistent contact depths, the variation in metrics illustrates that fusion is a process of presenting information and resisting interference. The standard deviation results of MS-SSIM show agreement with the global metrics, further indicating the fusion process effectively preserves structural features at each location.

This result demonstrates that image features can be effectively extracted and enhanced through our approach without compromising the integrity and reliability of the original information. Specifically, the wavelet transform allows us to process different frequency components independently and assign higher weights to critical regions for tactile perception. This approach is expected to enhance the tactile character and informativeness of the image, and the enhancement of the IE values is proof of this expectation. Meanwhile, the maintenance of the MS-SSIM values indicates that the fused image is comparable to the original image in terms of visual quality. The strategy enhances the tactile information without losing visual quality. The experiments have also shown that the number of images involved in fusion has an inconsistent correlation with the results in different scenes. But basically, it does not change the level of the metrics. Therefore, the number of samples required for performing fusion depends on the scene in the actual task.

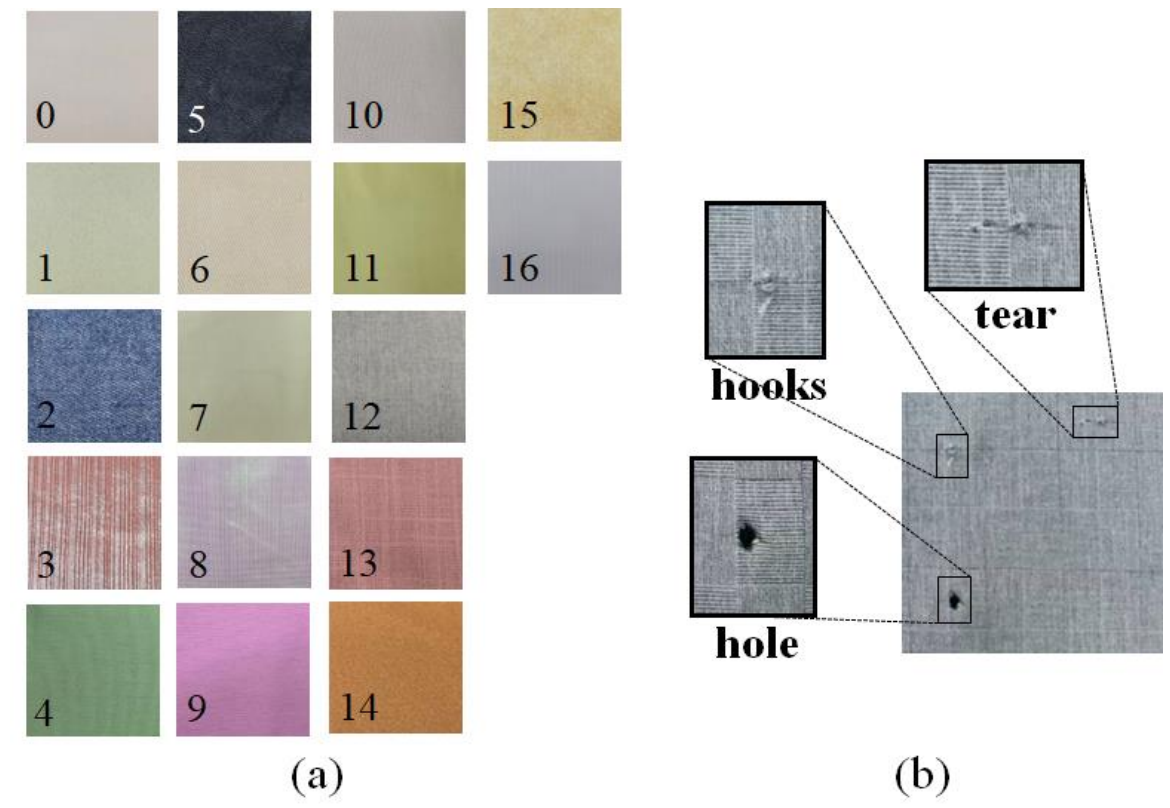

**Fig. 8.** (a)(b) Samples used to produce the dataset, include 17 fabrics and 3 defects.

## IV. Strategy application and expansion discussion

### A. Tactile image fusion of objects

Experiments show the process and results of performing contact information acquisition on daily objects. As shown in Fig.7(a), the sensor is connected to a robotic arm that can make contact with the object to be acquired, and the acquired image sequence is fused by our method. The figure shows the process of detecting a defective area of a piece of fabric, the captured raw image, and the result of remapping based on the distribution of the fused data. Fig.7(b) shows the effect on everyday objects such as screws, paper clips, and other objects. The image after fusion is performed can present a larger area of the object and more details.

### B. Fabric Detection Combined with Neural Networks

Features in an image that reflect information about an object can be extracted and analyzed, and specific targets can be identified and detected. Areas of a tactile image that produce significant contact can be utilized for object detection. The use of fusion strategies can enhance the ability to utilize information for detection. A fabric classification and detection scenario was used to conduct the tests, and this scenario represents a typical pattern for defect detection [25][26]. We discuss it through a test using neural networks[27][28], and a CenterNet neural network [29] using ResNet [30] as the backbone of the feature extraction network developed. Meanwhile, for the output structure of the network, an output branch was added for overall category classification to enable the network to take on both target detection [31] and category classification [32] tasks. We performed experiments on fabric samples. As shown in Fig.8(a) and Fig.8(b), 17 fabric samples were used to test the classification capability, and 3 different defect types were used to test the target detection capability. Vision-based tactile sensing system was used to touch and acquire images of the fabrics and label them with information. A discussion of neural networks is in Supplementary Material 4.

To validate and evaluate the value of circular fusion strategy among classification and detection. We selected a continuously collected sequence for evaluation. As shown in

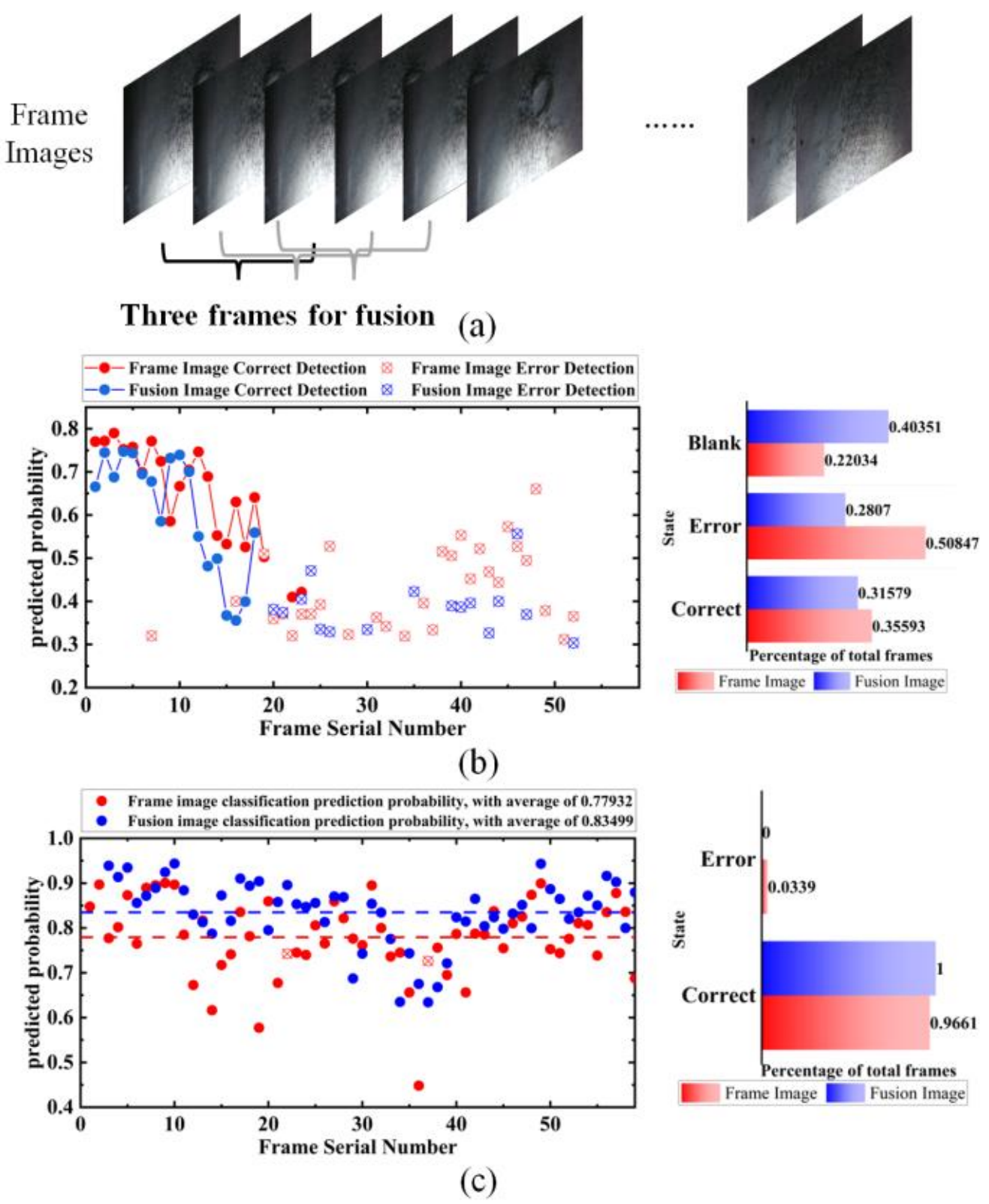

**Fig. 9.** (a) An image sequence for fabric defect detection. (b) Results of target detection performed on a sequence of frame images and fused images (c) Class prediction results on a sequence of frame images and fused images.

Fig. 9(a), a sequence of frame images containing 59 frames was sequentially merged into a fused image with 3 frame images. The frame images and the fused image were used as test data for prediction individually. Fig. 9(b) shows the variation and statistics of the prediction probability for defective target detection on the sequence, where the fused image reduces the probability of prediction error by 20% compared to the frame image without fusion. It can be noticed that the fused images behave more conservatively in the algorithm, meaning that they are more often assumed to be free of defects and are not predicted incorrectly. The reason for this phenomenon is the fusion process which makes some of the local features that may cause false interpretations to be weakened, thus reducing the probability of error, and in parallel with this some of the weak features that are used to determine the type of defects are also weakened in the process thus resulting in a leakage of detection. This skewing of the analysis of the data is uniquely valuable in scenarios where the accuracy of the prediction results is high and can reduce the generation of false predictions. Fig. 9(c) illustrates the predicted probabilities on fabric classification. The fused image achieves completely correct judgment and the prediction probability is higher than the frame image. This represents the extent to which the fusion algorithm enhances the global features, resulting in more accurate category prediction based on the comprehensive features of the image.

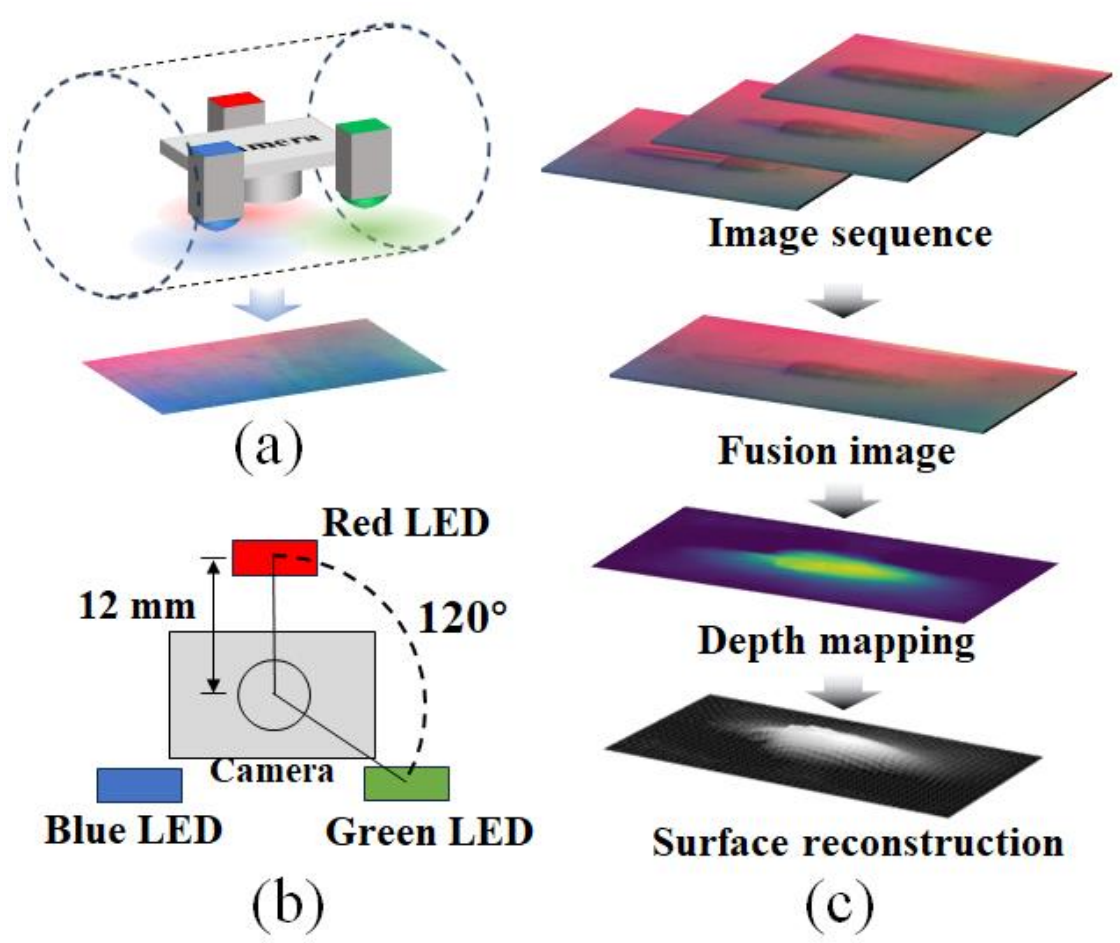

**Fig. 10.** (a) Source structure and raw RGB image for surface depth reconstruction. (b) The position of the three color light sources relative to the camera performing the imaging in the system. (c) Schematic of the use of the circular fusion strategy for surface depth reconstruction.

For complex classification tasks on surface continuous detection scenarios, the addition of fusion helps to improve prediction. More comparative analysis and visualization are presented in Supplementary Material 5, Video 1, and Video 2.

### *C. Reconstruction of surface depth*

Fusion of the images themselves is sometimes limited when acquiring richer contact states, and performing surface depth reconstruction can be more valuable. Surface depth reconstruction based on tactile images can present the state of a contact object in a comprehensive and highly accurate manner [11][33][34] and guide complex operations. Integrating the cyclic fusion strategy minimizes depth anomalies caused by surface contact during reconstruction, resulting in a more realistic restoration of the object's surface.

To ensure sufficient data for depth reconstruction, the sensing system incorporates red, green, and blue LED light sources, as illustrated in Fig. 10(a). These light sources are arranged in an equilateral triangle around the camera, as shown in Fig. 10(b), with each color independently received by the CMOS's three-channel receptors to form an RGB image. Before reconstruction, the cyclic fusion strategy processes the image information, as depicted in Fig. 10(c). Since fusion is processed for one value per pixel point, the RGB channels are separated for fusion and then combined into a color image. To achieve better imaging results, aluminum film is used in the system instead of the original reflective film. Details of the sensor changes are discussed in Supplementary Material 6.

To evaluate the contribution of the proposed strategy in the reconstruction, three 4mm diameter spheres are arranged to make contact with the surface. Ideally, the projected area of the spheres on the image after contact should match the projected area of the reconstructed spheres. The overlap of the areas can be used to evaluate the effectiveness of the reconstruction, as shown in Fig. 11(a)(c). Due to the presence of curved surfaces, the amount of contact produced by the three spheres is not equal. Three frame images from the continuously sampled image data performed for the spheres are used to perform fusion, as shown in Fig. 11(b). The response regions produced by the spheres on the image are numbered region 1 to Region 3, and Fig. 11(d) shows how the reconstruction of the projected areas of a single intermediate contact image and the fused image overlap, which represents the extent of reconstruction restoration. The results demonstrate that the image with the fusion strategy matches the three spheres more consistently than the image without the strategy, which is in line with the ideal situation. Further, the degree of matching at the reconstruction height also evaluates the effectiveness of the reconstruction. Due to the homogeneity of the sphere, the deepest part of the contact on each region was compared, as shown in Fig. 11(e)(f). The use of the fusion strategy resulted in the amount of variance in the regions being minimized.

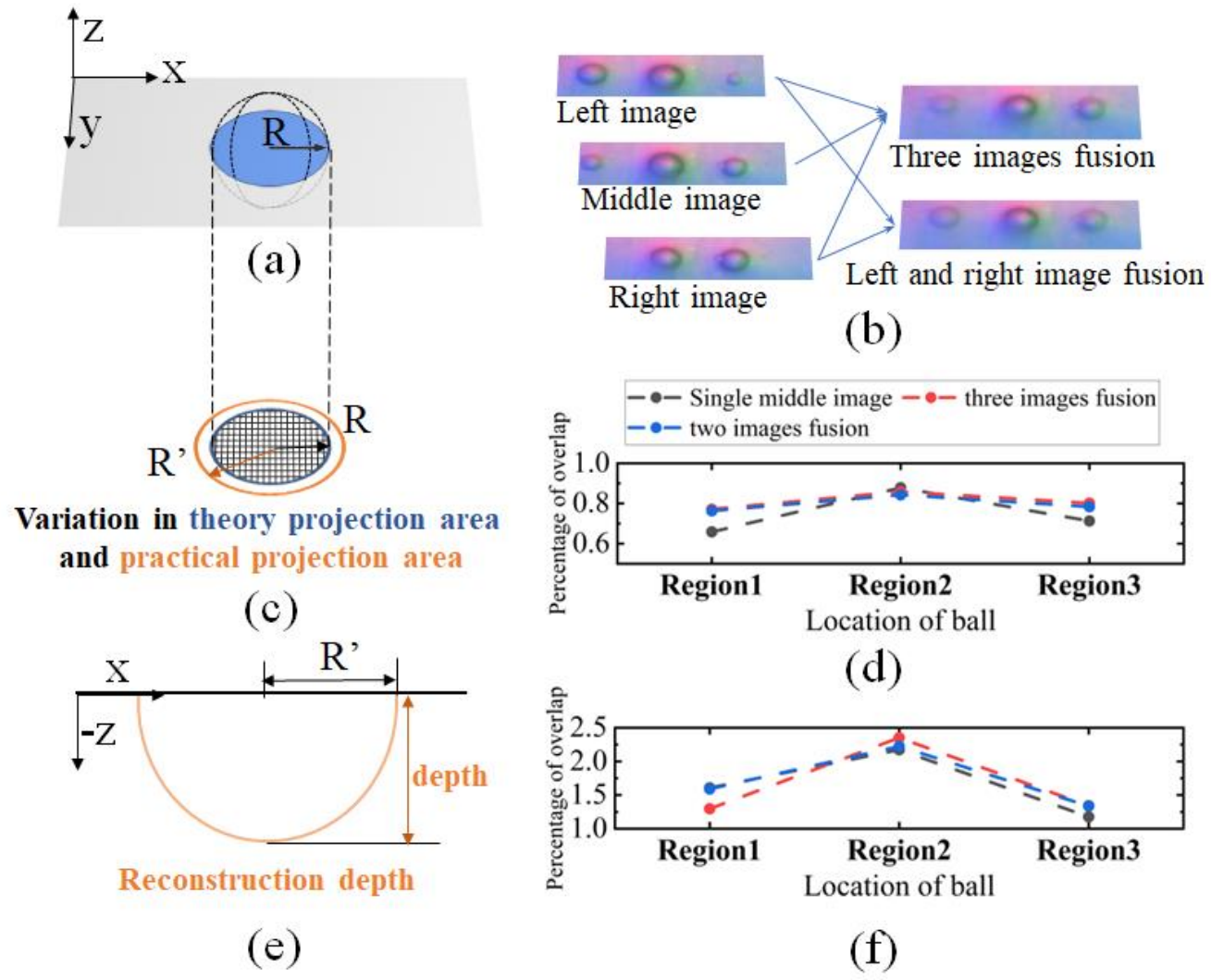

**Fig. 11.** (a) The projective plane of a sphere. (b) The tactile image obtained for the contact of the three spheres and the image obtained by performing fusion. (c) Difference between the reconstruction result and the initial image on the projection plane. (d) The overlap between the reconstruction result and the projection surface of the initial image on the response region. (e) The reconstructed height of the spheres. (f) The reconstruction height on the response region.

The reconstruction of the three equal-height spheres should ideally be uniform, but the curved surfaces introduce variations. The standard deviation of the response regions, calculated to quantify these deviations, is presented in Table IV. The application of the fusion strategy improved both metrics, reducing inconsistencies in contact depths across different image locations caused by the curved surface structure. Notably, a 50% reduction in offset on the projected surface. Further analysis of the number of images fused shows that fusing only the left image and the right image (which contact the left and right spheres) has better results than fusing three images. This occurs because each fusion weakens information unique to a single image, and additional fusion steps amplify this weakening effect, introducing further errors. This phenomenon is also reflected in the numerical changes of the reconstruction, and the numerical decrease due to fusion is demonstrated on the curves of Fig. 11(d)(f). Consequently, the number of fusions should be determined based on the specific scene.

TABLE IV

RESULTS OF REGIONAL DIFFERENCES IN RECONSTRUCTION

| Metrics | Single middle image | Three images fusion | Two images fusion |
| --- | --- | --- | --- |
| Standard deviation of overlapping areas of the projected area | 0.114221 | 0.044237 | 0.041682 |
| Standard deviation of reconstruction height | 0.500130 | 0.596887 | 0.457099 |

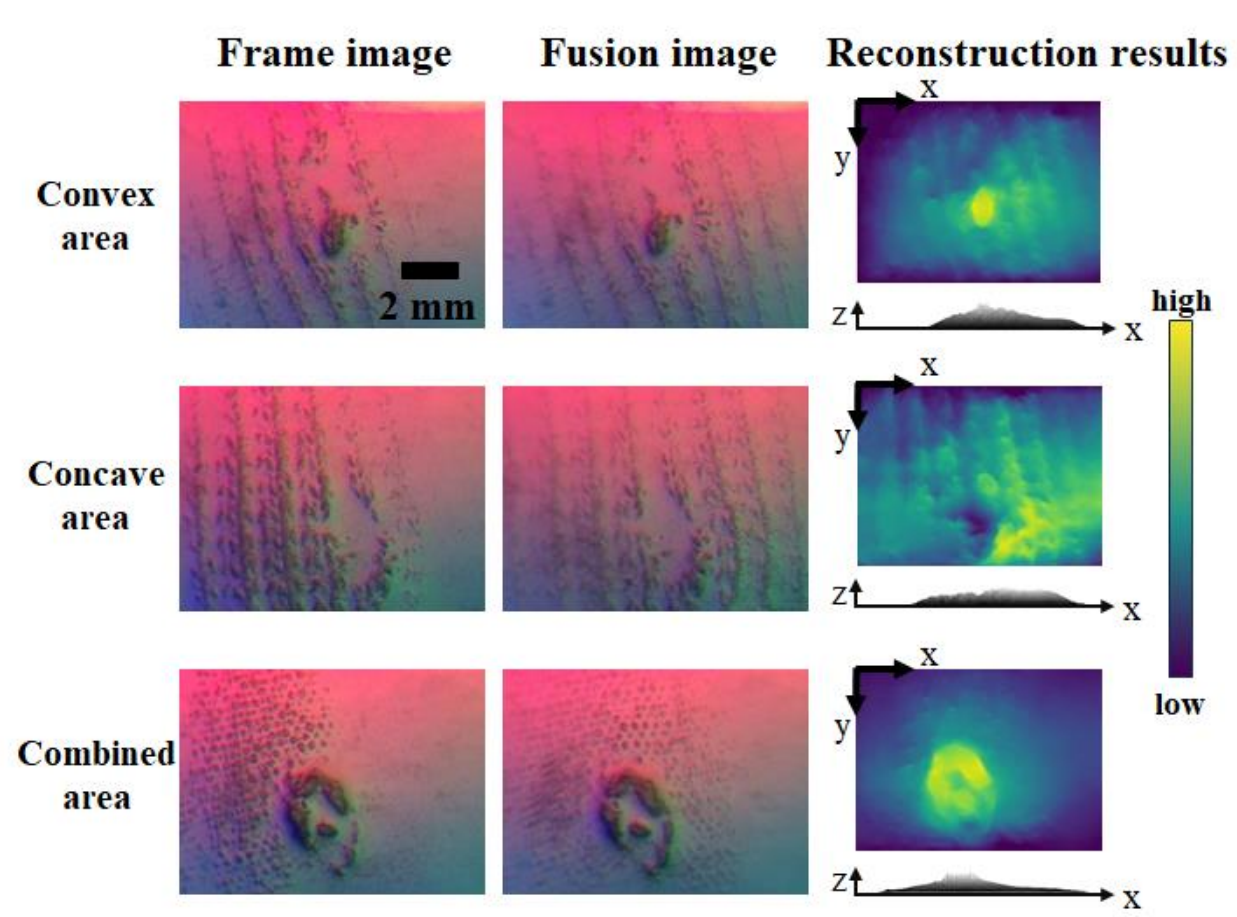

**Fig. 12.** The reconstruction results of single-frame images, fused images, and 2D displays obtained in the reconstruction of convex, concave, and combined regions of fabric defects.

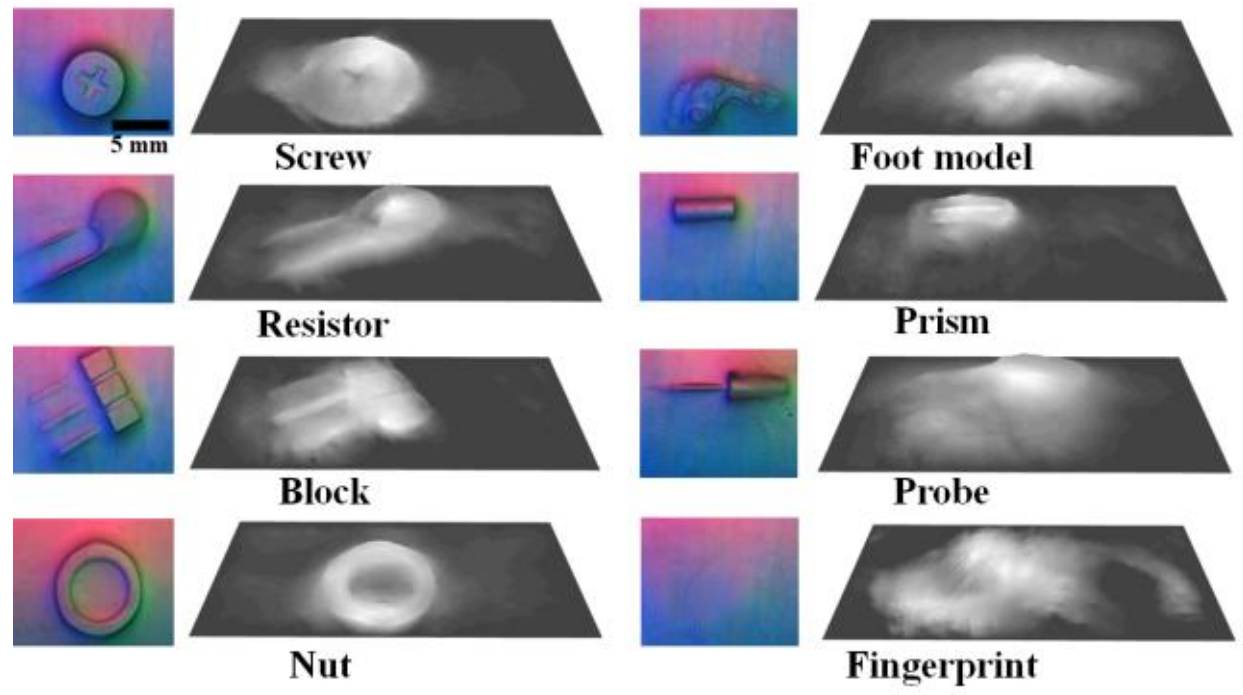

**Fig. 13.** Reconstruction of daily objects.

Effective surface reconstruction requires sensitivity to object convexity and concavity. Fig. 12 illustrates the calculation of convex, concave, and combined areas in reconstructing fabric defects. Fusion enlarges the regions where textures are rendered, enabling the reconstruction to accurately identify defect locations. Examples of reconstructing daily objects are shown in Fig. 13 and Supplementary Video 3. While the method effectively captures basic contours, it struggles with fine textures or regions with significant height variations, which are influenced by the elastomeric shape variables and light occlusion. Improving the sensor's contact layer and optimizing the light source design will be key areas for future work.

## V. CONCLUSION

In this work, we propose a cyclic fusion strategy to fusion and present actual elastomer deformations in curved surface contact. For the situation that surface contact data are at different depths due to the curved surface contact structure, we use cyclic fusion to extract and integrate the data that characterize the object features by combining the characteristics of the contact depths and the changing properties of the object features in the image. A cylindrical vision-based tactile sensing system with typical curved contact characteristics is fabricated for testing and discussion. The cyclic fusion is performed by cyclically performing wavelet transform followed by averaging fusion and fusion based on significance analysis for the low-frequency and high-frequency components. The tactile images fused with our proposed strategy have over 8% more information and effectively reduced 20% for regional information bias while similar structural similarity as compared to the single images captured by the sensing system and the images obtained using the sampling stitching method. In neural network-based defect detection applications, the use of this fusion approach led to a 20% reduction in erroneous predictions. Furthermore, when applied to surface depth reconstruction, the strategy significantly reduced the projection bias resulting from surface curvature by 50%. Our approach both overcomes the lack of non-uniform depth present in curved contact structures and extends the study with potential applications utilizing structural features. Combining the immediacy of single images with the comprehensiveness of fused images, we foresee a new multimodal sensing paradigm with feasibility. The sensor needs to scan the object only once to obtain an exhaustive data set. This approach not only improves detection efficiency and reduces repetitive work but also enhances the recognition of object features by utilizing a single image's high resolution and detail information while maintaining the continuity of haptic perception. Future improvements in our work include integrating the sensor hardware and the algorithm's robustness.